# MindBigData 2023 MNIST-8B

## The 8 billion datapoints Multimodal Dataset of Brain Signals

**David Vivancos**

Email: *vivancos@vivancos.com*

**Abstract**

*MindBigData 2023 MNIST-8B is the largest, to date (June 1st 2023), brain signals open dataset created for Machine Learning, based on EEG signals from a single subject captured using a custom 128 channels device, replicating the full 70,000 digits from Yaan LeCun et all MNIST dataset. The brain signals were captured while the subject was watching the pixels of the original digits one by one on a screen and listening at the same time to the spoken number 0 to 9 from the real label. The data, collection procedures, hardware and software created are described in detail, background extra information and other related datasets can be found at our previous paper "MindBigData 2022: A Large Dataset of Brain Signals".*

## 1.- Introduction

One of the main metrics used over the last decade to measure the performance of machine learning and deep learning algorithms, has been the MNIST [1] dataset, as of June 2023 the accuracies obtained are at 99,87% [2], it is fair to say that peak performance is already achieved, so I think it's the time to add a new layer of complexity, and for that, what a better way than recreating the dataset, but with brain signals captured while watching the real pixels of the original MNIST digits and listening at the labels at the same time.

And with this increasing the datapoints from almost 72 million in the original dataset to more than 8 billion at MindBigData 2023 MNIST-8B, so an increase of more than 100X, with the same labels but now with several possible objectives, from the classification of the 10 labels, to the reconstruction of the 70,000 digits pixels or the 10 audio waves listened based on the brain signals as inputs, to many other possibilities, yet to imagine and explore.

With the release of this open dataset the aim is to two-fold, **increase our knowledge of the brain** and have a **new challenging metric for the Machine Learning community** to foster the development of new algorithms to beat it, and advance the Artificial Intelligence field.

## 2.- Properties

The dataset contains 140,000 records from 128 EEG channels, each of 2 seconds, recorded at 250hz, in total 17,920,000 brain signals and 8,960,000,000 data points.

The dataset is hosted at HuggingFace:

https://huggingface.co/datasets/DavidVivancos/MindBigData2023_MNIST-8B

It consists of **2 main data files**, described below:

"train.csv"
"test.csv"

**10 audio files** at a folder named "audiolabels":

"0.wav", "1.wav"…. "9.wav"

And 1 file with **3d coordinates** of the EEG electrodes:

"3Dcoords.csv"

The brain nomenclature for the EEG [3] channels used is based on a selection from the 10-5 system [4] with the following distribution:

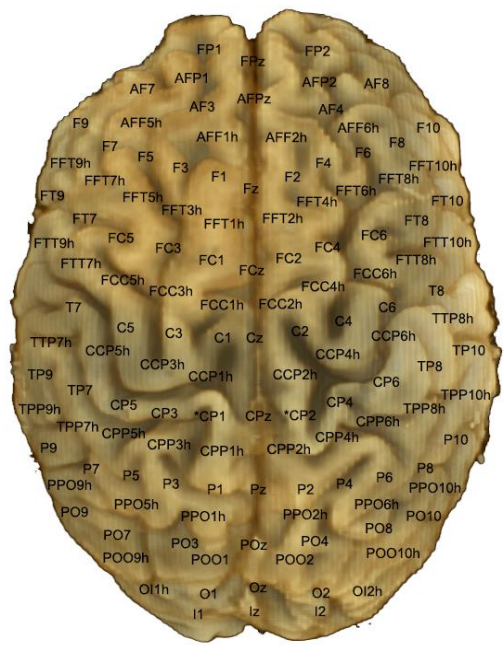

So the 128 EEG Channels used are: FP1, FPz, FP2, AFp1, AFPz, AFp2, AF7, AF3, AF4, AF8, AFF5h, AFF1h, AFF2h, AFF6h, F9, F7, F5, F3, F1, Fz, F2, F4, F6, F8, F10, FFT9h, FFT7h, FFC5h, FFC3h, FFC1h, FFC2h, FFC4h, FFC6h, FFT8h, FFT10h, FT9, FT7, FC5, FC3, FC1, FCz, FC2, FC4, FC6, FT8, FT10, FTT9h, FTT7h, FCC5h, FCC3h, FCC1h, FCC2h, FCC4h, FCC6h, FTT8h, FTT10h, T7, C5, C3, C1, Cz, C2, C4, C6, T8, TTP7h, CCP5h, CCP3h, CCP1h, CCP2h, CCP4h, CCP6h, TTP8h, TP9, TP7, CP5, CP3, Cpz, CP4, CP6, TP8, TP10, TPP9h, TPP7h, CPP5h, CPP3h, CPP1h, CPP2h, CPP4h, CPP6h, TPP8h, TPP10h, P9, P7, P5, P3, P1, Pz, P2, P4, P6, P8, P10, PPO9h, PPO5h, PPO1h, PPO2h, PPO6h, PPO10h, PO9, PO7, PO3, POz, PO4, PO8, PO10, POO9h, POO1, POO2, POO10h, O1, Oz, O2, OI1h, OI2h, I1, Iz and I2.

CP1 was used as reference and CP2 as ground.

## 2.1.- Data Files

The 2 main data files are in CSV format [5], and structure is the same for both, a header row and then 60,000 data rows for the train file, and 10,000 data rows in the case of the test file. All the fields are separated by the comma character "," .

Each data row includes the EEG data, label, and extra information described below, each for a single digit captured for 2 seconds or a "black" blank space of 2 seconds in between.

The structure of each row is as follows:

The first **64,000 fields hold the raw EEG** data captured from the sensors without any further pos-processing, as it comes from the amplifiers and ADCs [6] used, since there are 128 channels recorded at 250hz of 2 seconds each, we have 64,000 data points per row (250 x 2) x 128.

The data type for all these fields is an integer numeric value.

The order follows the channel listing above, first 500 samples (250hz x 2 seconds) for the channel "FP1", then 500 samples for "FPz", then 500 for "FP2"….and so on for all the channels, and lastly 500 samples for the channel "I2".

The headers reflect this with the channel name followed by an underscore character "_" and the sample number, or "time" order, from 0 to 499 for each channel like:

FP1_0,FP1_1,FP1_2,FP1_3,FP1_4,………..FP1_499 and so on for the 128 channels ending with column 64,000 being "I2_499"

Then the column 64,001 will be "**label**" referring to the original MNIST label, could be any number from 0 to 9, also could be "-1" to indicate the blank space captured before the digits.

Followed by the column 64,002 that will be "**label_source**" referring to the source of the original MNIST label, can be the text "TRAIN" or "TEST".

Then the column 64,003 will be "**label_pos**" referring to the location in the original MNIST dataset, can be a number from 0 to 59,999 if the previous field is "TRAIN" or from 0 to 9,999 if the previous field is "TEST".

Note that "label_source" and "label_pos" may not be needed but were added to include extra context information referring to the original MNIST dataset.

After these 3 label fields we have 784 values named "**label_imgpix**_0","label_imgpix_1","label_imgpix_ 2"… up to "label_imgpix_783". The field holds integer values, the possible values range from 0 to 255 like in the original MNIST digits, related to the pixels intensities in gray scale, using these values it is possible to create the 28x28 pixels image that is shown in the screen (amplified) as the stimulus for the brain waves, like for example the handwritten digit 9 below:

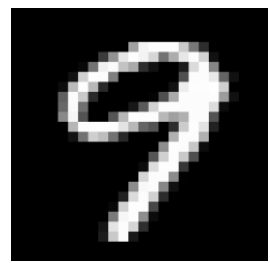

For the label "-1" or the blank 2 seconds space in between the digits a full black screen is shown so all the 784 "label_imgpix_" values will be 0s.

After these 784 values there are 4 ending fields provided also as extra information:

"**timestamp**" with a Unix Like timestamp long integer value referring the starting date/time of capture for each row.

"**sessionnum**" a numeric value referring the chronological order of the session where this capture was done, refer to the chapter 3.3 for more information on this regard.

"**blocknum**" a numeric value referring to the number of block (usually of ten captures) where this capture was done, refer to the chapter 3.3 for more information on this regard.

"**blockpos**" a numeric value with the position in each block of captures, where this specific capture was done, refer to the chapter 3.3 for more information on this regard.

## 2.2.- Audio Files

For the 0 to 9 digits, once it was show on the screen, an audio file was played at the same time, with the corresponding label spoken, note that the duration of the audio files was shorter than the full 2 seconds that the digit was shown on the screen.

The audio files are single channel mono WAV [7] files at 24,000hz 16 bits per sample, PCM format, with the following duration/samples:

"0.wav" 321 milliseconds with 7,713 samples.

"1.wav" 450 milliseconds with 10,817 samples.

"2.wav" 510 milliseconds with 12,257 samples.

"3.wav" 440 milliseconds with 10,561 samples.

"4.wav" 262 milliseconds with 6,305 samples.

"5.wav" 301 milliseconds with 7,233 samples.

"6.wav" 454 milliseconds with 10,913 samples.

"7.wav" 270 milliseconds with 6,497 samples.

"8.wav" 224 milliseconds with 5,377 samples.

"9.wav" 364 milliseconds with 8,737 samples.

## 2.3.- 3D coordinates file

For possible advanced exploration of the EEG sources in the brain anatomy, source localization or brain connectivity applications, 3d locations of each sensor is provided, it was generated using a video covering the shape of head wearing the built EEG cap, and then using Nerf [8] algorithms to recreate the 3d geometry.

Once the geometry was generated, the center coordinates (0,0,0) of the scene was placed in a middle point inside the skull, and then all the 128+2 sensor location (x,y,z) coordinates were annotated (selecting a point in the middle outside of each sensor) to produce the coordinates file:

"3Dcoords.csv"

The file is a CSV file with 4 columns, a header row and 130 data rows. All the fields are separated by the comma character ",", and the point "." character is used as the decimal point in the coordinates values.

The First column is the "channel" name, followed by "x","y" & "z" coordinates.

All the data rows are like the following example:

"FP1,-0.089225,0.162751,0.206887"

Notice that 2 extra channels coordinates were included at the end, CP1 (reference) and CP2 (ground), even if there is no direct EEG data captured for these channels, could potentially be useful for some re-referencing possible scenarios.

## 3- Constructing MindBigData 2023 MNIST-8B

MindBigData [9] started in early 2014, initially capturing brain signals with several consumer EEG devices, explored deeply in the paper "MindBigData 2022: A Large Dataset of Brain Signals" [10].

But it was clear that new devices were needed to due to the complexity and wide distribution of brain activities.

With that in mind, the goal was to use higher density EEG devices to extend the scope, the problem with it was that high density EEG devices with 64, 128 or 256 channels are orders of magnitude more expensive compared to commercial non-medical grade wireless EEG devices [11] used so far with only up to 14 channels. And the setup to start capturing in high-density EEG medical grade are usually more complex, involving more time to get them ready and clean up after the captures.

The first step was to **try to reduce the cost by building our own devices**, starting in 2021 with a custom 64 channels device, that was used in 2022 to capture a very small subset also of the MNIST dataset, it was created by combining 2 x 32 EEG channels medical amplifiers, building a custom cap, and creating the software to sync the devices and generate the 64 channels output.

The problem encountered with that device was that the reliability of the signals, was not stable and good enough to capture at sufficient speed, so it was not feasible to capture a large enough dataset in a reasonable timeframe, and capture with this device stopped at the end of 2022.

So, in 2022 I decided that the goal should be to build every component involved in an EEG device, that way all the process can be controlled measured and improved, also increasing the number of channels to 128 to have a better coverage of brain anatomy [12]. And reducing the cost of a commercial device at least one order of magnitude.

The device construction and testing to capture MindBigData MNIST-8B was finished by the end of 2022.

## 3.1- Main components

There are several components when building a noninvasive EEG system or more broadly a BCI [13] Brain Computer Interface:

1.- The sensors or brain activity measuring devices, and the holding mechanism to the head.

2.- The connectivity, from the sensors to the other parts, can be wired or wireless if the bandwidth allows, for wired there is usually a "brain box" where all the cables coming from the head end.

3.- The ADC and signal amplifier/s to boost the faint electrical signals captured by the sensors from the brain waves and converting them into a digital format, also this device oversees sending the data to a computer system where it can be stored/processed.

4.- The Firmware that deals with the 3 previous components and finally the Software that deals with the storage and processing of the raw data captured.

With all that in mind, after wide research over all the existing initiatives in the open hardware/software landscape and other commercial options, the selection was clear:

1.- (130) Dry passive Silver/Silver Chloride **electrode sensors** [14] were selected, with 12mm diameter and several prong heights to cover different areas of the head morphology, so they can pass through hair, and with a DC resistance ≤ 500 Ω.

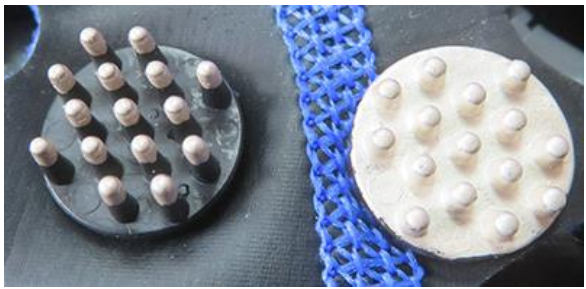

Dry electrodes were selected, to increase the wearability and reduce the setup time, with some tradeoff regarding the increased impedance versus some wet gel alternative electrodes, or even active electrodes.

A pre-assigned 130 mounting holes 10/5 EEG system flexible **textile cap** was used to hold the electrodes.

2.- For the **connectivity**, wireless was discarded due to the data throughput wanted, and to simplify the on-head components, so several cabling options were reviewed and tested, including shielded cables, but finally due to weight and flexibility constrains a **custom cable** was designed and built with a very small diameter of 0.6mm and produced with a standard 4mm snap connector on one side a standard din 1.5mm on the other side.

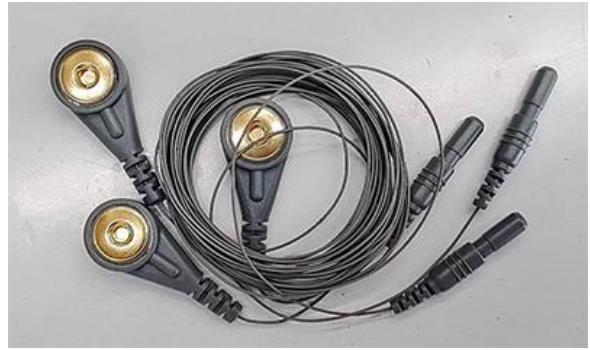

They were connected to a custom brainbox for organization and selection of the channels, and from there to the "Controlling device" or the ADC/Amplifier.

The following image shows a snapshot of the final cap, with the electrodes and cabling.

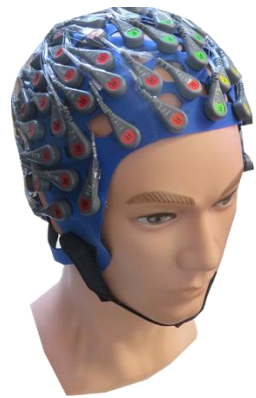

3.- Regarding **the "Controlling device"** or the ADC and signal amplifier, without any doubt the more complex component, after long research in the subject, was clear that the more robust one was the open-hardware approach by NeuroIDSS [15] available at Github, and already with an alpha prototype version named **FreeEEG128**, with hardware schematics blueprints and basic firmware source code provided.

The only caveat is that was and still is in "alpha" state of development, and to date probably less than 5 units of this device have been built worldwide, being one of them, the one I built for MindBigData, so it was a fully exploratory process.

Even so with the Gerber files [16] and bill of materials, the device was built, with only a few modifications due to some component's shortages in 2022, and once the 130+ cables were produced, a few extra connector pieces produced and many soldering the device was ready to test and move to the next step.

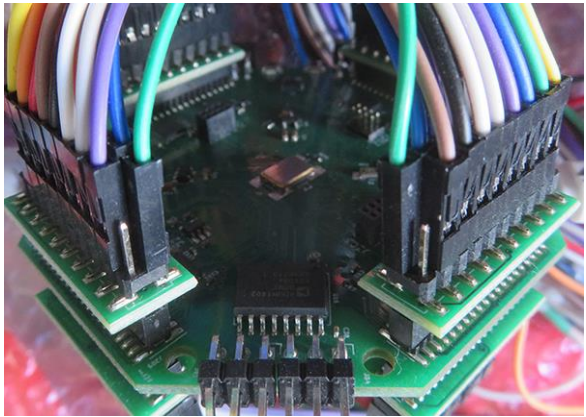

One of the critical components of the hardware is the ADC, this device uses 16 Texas Instruments ADS131m08 [17], one for every 8 EEG channels, since this in essence also a differential amplifier, I arbitrarily selected for each of the amplifiers 8 as separate as possible EEG channels on head, to increase possible signal singularity.

If this is of interest for further research, here is the 16 ADCs, distribution each with the 8 EEG channels connected (CP1 is reference and CP2 ground for all):

| ADC1 | F9 | P9 | C3 | Fz | Pz | C4 | F10 | P10 |
|---|---|---|---|---|---|---|---|---|
| ADC2 | FT9 | TP9 | AFF5h | POO9h | AFF6h | POO10h | FT10 | TP10 |
| ADC3 | T7 | FFC5h | CPP5h | FPz | Oz | FFC6h | CCP6h | T8 |
| ADC4 | C5 | AF7 | PO9 | FCC1h | CPP2h | F8 | C6 | PO8 |
| ADC5 | F7 | PPO9h | AFF1h | CPP1h | FCC4h | F6 | OI2h | PPO10h |
| ADC6 | TPP9h | FC5 | CPP3h | Cz | AF4 | FC6 | CPP4h | TPP10h |
| ADC7 | FFT9h | P7 | FP1 | C1 | F2 | I1 | TTP8h | PPO6h |
| ADC8 | FFT7h | TPP7h | CCP1h | FP2 | FFT10h | FCC6h | P8 | POO2 |
| ADC9 | FTT9h | P5 | FFC1h | PPO2h | AF8 | FTT10h | P6 | PO10 |
| ADC10 | F5 | CP5 | AFp2 | Cpz | FC4 | CP6 | PO3 | I2 |
| ADC11 | F3 | CCP3h | TP7 | PPO5h | CCP4h | PO4 | FFT8h | TP8 |
| ADC12 | TTP7h | AFp1 | FC1 | O1 | F4 | C2 | P4 | FTT8h |
| ADC13 | FT7 | CCP5h | AF3 | PPO1h | OI1h | FC2 | FT8 | TPP8h |
| ADC14 | FTT7h | FC3 | PO7 | P1 | AFF2h | CCP2h | CPP6h | O2 |
| ADC15 | FCC5h | CP3 | FFC3h | FCz | FFC4h | CP4 | P2 | POz |
| ADC16 | F1 | FCC3h | Iz | P3 | FFC2h | FCC2h | AFPz | POO1 |

4.- For the **Firmware**, a basic open-source code was provided for the operation of the STM [18] RISC processor in the device, only choosing and coding for example the ADCs frequency (250hz) or samples per second, gain (32) and other parameters, so once it was compiled and flashed to the device it should be ready to start testing and sending data via USB to the host machine.

### 3.2- The Software

Also, the needed software was built to dealt with the raw data captured from the EEG device, it only sends raw packets generated at the FreeEEG128 device using a simulated serial port over USB.

The raw EEG are 24 bits for each sample and channel coming out of the 16 TI ADS131m08 configured at gain 32 mode, being vref 1.25V, so the units are not "real" EEG microvolts, for that a calculation with these raw numbers should be performed, something in the lines of:

raw * (1250000 / ((pow(2., 23) - 1) * 32))

But I decided to provide data in raw format only, to avoid pre-processing and because to have a number closer to "real" EEG microvolts, for example "half-cell potential" [19] should be taken into consideration, and that depends on calculation of material properties of the sensors and other factors, beyond the scope of this research and dataset.

As the **data is provided in raw format**, it can be used as so, since for some algorithms maybe suitable, or some pre-processing is usually performed when doing EEG analysis [20], like **removing the 50hz line noise** if it is present (capture was done in San Lorenzo de El Escorial, Spain, Europe, so 50hz maybe present), a **baseline correction** [21] with maybe a FIR high-pass filter at 0.1Hz, and a **Band pass filter** usually between 0.1Hz and 40 or 50Hz too, but my advice is to also explore beyond what is "usual", EEG is inherently very noisy [22], but maybe there is more in the signal that gets altered due to the filtering, future explorations will tell.

Once the data flow from the EEG Device to the host machine starts, it is stored in real time for each channel including a recording timestamp for each packet of 128 channel data. 250 packets are captured and stored in a buffer in memory per second in "real time".

With that packets some functions were built to visualize the raw waves for each channel and a custom dashboard to measure a quality metric in real time for each channel, using a PSD algorithm [23].

The **interactive dashboard** is shown in screenshot below from red meaning flat sensor or no data (could be a floating sensor or some other factor impacting the impedance) to green meaning in principle good signal quality.

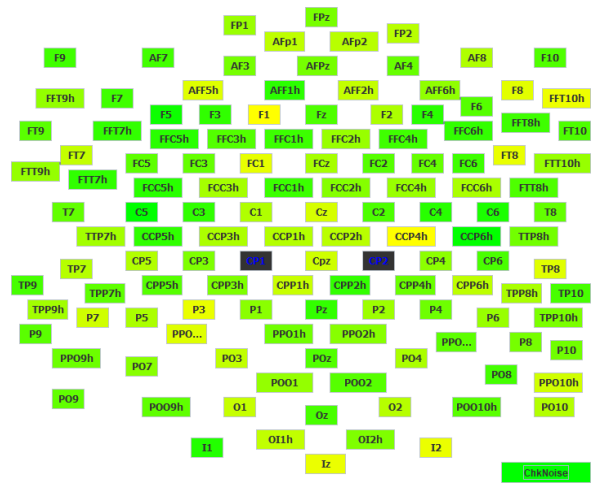

The idea was to review it before starting each capture and be sure that most or all the channels have a good enough signal quality, since the capture of MindBigData2023 MNIST-8B involved hundreds of hours of capture, it was decided to start the capture as long as 96 or 97% of the channels were with good or medium quality signals, meaning no more than 4 or 5 "dead" or flat channels.

In high density EEG devices, discarding some channels is usual, due to several factors, will be great to have all of them always at good quality, but here a tradeoff was needed to meet the expected time scheduled to capture the 140,000 of 2 seconds each.

Also, a custom software was built to display the MNIST digit pixels, like for example the following digit "2":

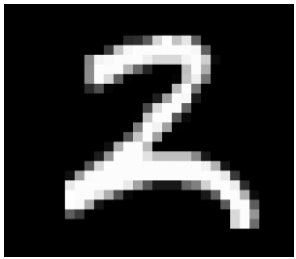

And code to play the audio waves at the same time the digit is shown, and to store the data and labels.

### 3.3- The Capture Procedure

The capture procedure is very similar to previous MindBigData datasets.

The subject, David Vivancos [24] (author of this paper) is always seated, in the same environment for each recording, in front of a 65" screen that is connected to a computer where the custom capturing software is running.

Once the Controlling device is powered via an USB power bank (to reduce line noise), the EEG sensors cap is placed on head, and a manual check is done in each sensor to ensure they are touching the scalp, the USB cable is connected to the computer, then the custom capturing software is started, once it properly connects to the device, just a signal quality check is performed as mentioned in chapter 3.2.

If quality is good enough, signal capture can start, it is done in **sessions** (look for "sessionnum" field in the data files to find out in what session each capture was done) hundreds of them where needed to record the brain signals for the 70,000 MNIST digits and the 70,000 black spaces in between.

**Session duration was not fixed**, since it depended on keeping a good signal quality, not too high room temperature since can induce sweating and distort the channels, measured attention level, and other factors, so could be a few minutes to up to 1 hour, including up to 500 or more digits captured in a single session.

Each session was divided into **blocks or batches** to avoid fatigue and to leave time for rest in between if needed, (look for "blocknum" field in the data files to find out in what block each capture was done, block number start from 0 each session up to 100 or more depending on session duration).

Each block usually includes **10 continuous captures**, interleaving 5 MNIST digits, and 5 black spaces in between, each of 2 seconds and **starting with a black space and ending with a digit**. (To locate the position, look for "blockpos" field in the data files to find out what is the exact position or order in the block where the digit or black space was captured, number starts from 0 each block up to 9, rarely less if block capture needed to stop earlier for some reason)

Have in mind that recording of brain signals is done in a fully **controlled experiment**, carefully trying to avoid, as much as possible, any body part movements or blinks [25]. And this is the reason to do the capture is small blocks of 10 captures (5 digits + 5 blanks) or **20 seconds of full concentration per block**.

The start of the block is triggered manually by the subject, with the click of a button, leaving a couple seconds after pressing it, to have some time to setup before the first black space and digit capture starts.

Once the capture of the block (5 digits + 5 blanks) is finished, it is shown in the screen the number and locations of flat signal sensors, if any happened during the block, then the **subject can decide to save it** by clicking a button, **or maybe decide to not save**, and discard this block, in case there are many bad channels, any distraction happened, movement, noise, or for any other reason.

Failed block captures are not saved, the only way to infer, (but without full certainty) is looking at the timestamps between each capture or block capture, since block recording time + save time + setup time is usually constant, notice that the first few 1000s recordings the software function to save the EEG data to disk was not optimized enough, and took longer, this was fixed later on, so saving between blocks was improved to just a few seconds.

The digits show in each block follow the order of the original MNIST digits first the 60,000 train digits, followed by the 10,000 test digits. The subject besides having recorded some of them in the past with other EEG devices has no direct clue of what digit is going to be shown and listened.

Since each block is usually 5 digits, if the block is saved these 5 digits are removed from a pull of pending digits, so they are not shown again, but if the block is not saved, the next block will have the same 5 digits as the failed block, so in these scenarios memory could play some role, but failed and repeated blocks are rare, usually less than 5 in each session).

The capture of the full 70,000 original MNIST digits was finished on May 31st 2023. And all the EEG produced is made available with the release of this paper as Open Database License.

**4- Discussion and Future Work**

Worth noticing that over the course of 2023 several advances in the field have started to show incredible results in decoding brain activities to images [26] [27] or even text [28] but all of them used FMRI [29] datasets, and this technology is not suited to be used anywhere and equipment to capture them is very expensive, versus the EEG used here.

Finishing the recording of MindBigData 2023 MNIST-8B was a great milestone in the MindBigData agenda to provide data and algorithms to decode brain activities into actionable commands and advance the Neurotechnologies and Machine Learning fields.

For me the AI field (and others too) in general needs 3 things data + algorithms + hardware [30], and **open and big enough data was still a missing link for brain science**, and this is where this dataset fits.

The main reason to dedicate so much time building the devices, capturing the brain signals and making it open-data is simply to provide the data needed to everyone interested in getting involved in this challenging field.

I encourage also the experienced Machine Learning community to get also into the brain science field, **using this data as a benchmark to prove novel algorithms**, will be good to have also open online competitions, with representation, tutorials or panels on the main events of the field like ICML [31] , ICLR [32] , Neurips [33] or CVPR [34], and even start a discussion about Neuro-ethics too at responsible oriented events like United Nation's ITU AI4Good [35].

Once the algorithms, built by the community to decode the dataset, reach a high enough accuracy it will create a wide range of opportunities to keep exploring our brains and learning more about how they work.

Regarding the next steps, using the built 128 EEG Channels device, more datasets will be created with several other modalities and other capture resolutions will be explored.

Trying to decode trillions of neurons [36] firing with only 128 "reading" points in the surface of our head is a daunting task and looks like it has a clear limit imposed by the laws of physics.

So, the Journey could not end here, we still need better tools, so besides working also in novel algorithms, in 2023 started building what one day hopefully will be a **1024 EEG channels device to increase 10X the previous resolution**, with also an open approach, to reduce the cost as much as possible and extend our limits. The following image shows the almost 4,000 sensors parts already produced for a first prototype:

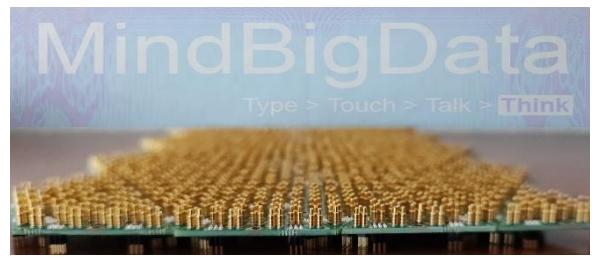

Stay tunned the next chapter, in the journey from Type to Touch to Talk to Think [37].

**5.- Conclusion**

A novel open massive dataset with more than 8 Billion datapoints is provided to foster the development of Machine Learning algorithms and improve the field of Neurotechnologies & Brain Computer Interfaces by having a new metric of brain signals decoded into 10 actionable commands.

**6.- Acknowledgment**

First, thanks to the 100+ researchers worldwide already using the existing open-data's at MindBigData, since the works they produce with this data really give meaning to the tedious, resource and time consuming needed to create them.

Also, thanks to the booming startups that challenge the limits of the Neuroscience and Neurotechnology field, from the pioneers to newcomers since there is much room to improve the field for all.

And of course, MindBigData 2023 MNIST-8B won't be possible, without the worldwide vibrant community of open-science builders, makers and explorers, that share experiences, discoveries and hurdles of trying to build hardware and software, for a human-device we are all born with, our brains, and one that we know so little off. So, thanks also to NeuroIDSS [38], NeuroTechX [39] and BrainsatPlay [40] and others.